\documentclass{article}

\usepackage{microtype}
\usepackage{graphicx}
\usepackage{subfigure}
\usepackage{booktabs} 
\usepackage[cjk]{kotex}
\usepackage{amssymb}
\usepackage{amsmath}
\usepackage{natbib}
\usepackage{multibib}
\usepackage{algorithm} 
\usepackage{algorithmic}  
\usepackage[titlenumbered,ruled,algo2e]{algorithm2e}
\allowdisplaybreaks
\usepackage{hyperref}
\usepackage[accepted]{icml2019}
\newcommand{\ie}{\textit{i}.\textit{e}.}

\icmltitlerunning{Learning Condensed and Aligned Features for Unsupervised Domain Adaptation Using Label Propagation}

\hypersetup{draft}

\begin{document}

\twocolumn[
\icmltitle{Learning Condensed and Aligned Features \\for Unsupervised Domain Adaptation Using Label Propagation}



\icmlsetsymbol{equal}{*}

\begin{icmlauthorlist}

\icmlauthor{Jaeyoon Yoo}{equal,to}
\icmlauthor{Changhwa Park}{equal,to}
\icmlauthor{Yongjun Hong}{to,goo}
\icmlauthor{Sungroh Yoon}{to}
\end{icmlauthorlist}

\icmlaffiliation{to}{Department of Electrical and Computer Engineering, Seoul National University, Seoul, Korea}
\icmlaffiliation{goo}{ENERZAi, Seoul, Korea}

\icmlcorrespondingauthor{Sungroh Yoon}{sryoon@snu.ac.kr}


\icmlkeywords{Machine Learning, ICML}

\vskip 0.3in
]



\printAffiliationsAndNotice{\icmlEqualContribution} 
\begin{abstract}
Unsupervised domain adaptation aiming to learn a specific task for one domain using another domain data, has emerged to address the labeling issue in supervised learning, especially because it is difficult to obtain massive amounts of labeled data in practice. 
The existing methods have succeeded by reducing the difference between the embedded features of both domains, but the performance is still unsatisfactory compared to the supervised learning scheme. 
This is attributable to the embedded features that lay around each other but do not align perfectly and establish clearly separable clusters. 
We propose a novel domain adaptation method based on label propagation and cycle consistency to let the clusters of the features from the two domains overlap exactly and become clear for high accuracy.
Specifically, we introduce cycle consistency to enforce the relationship between each cluster and exploit label propagation to achieve the association between the data from the perspective of the manifold structure instead of a one-to-one relation. 
Hence, we successfully formed aligned and discriminative clusters. 
We present the empirical results of our method for various domain adaptation scenarios and visualize the embedded features to prove that our method is critical for better domain adaptation.
\end{abstract}

\section{Introduction}

\begin{figure}
  \centering
  \begin{tabular}{@{}c@{}}
    \includegraphics[width=\linewidth,height=5cm]{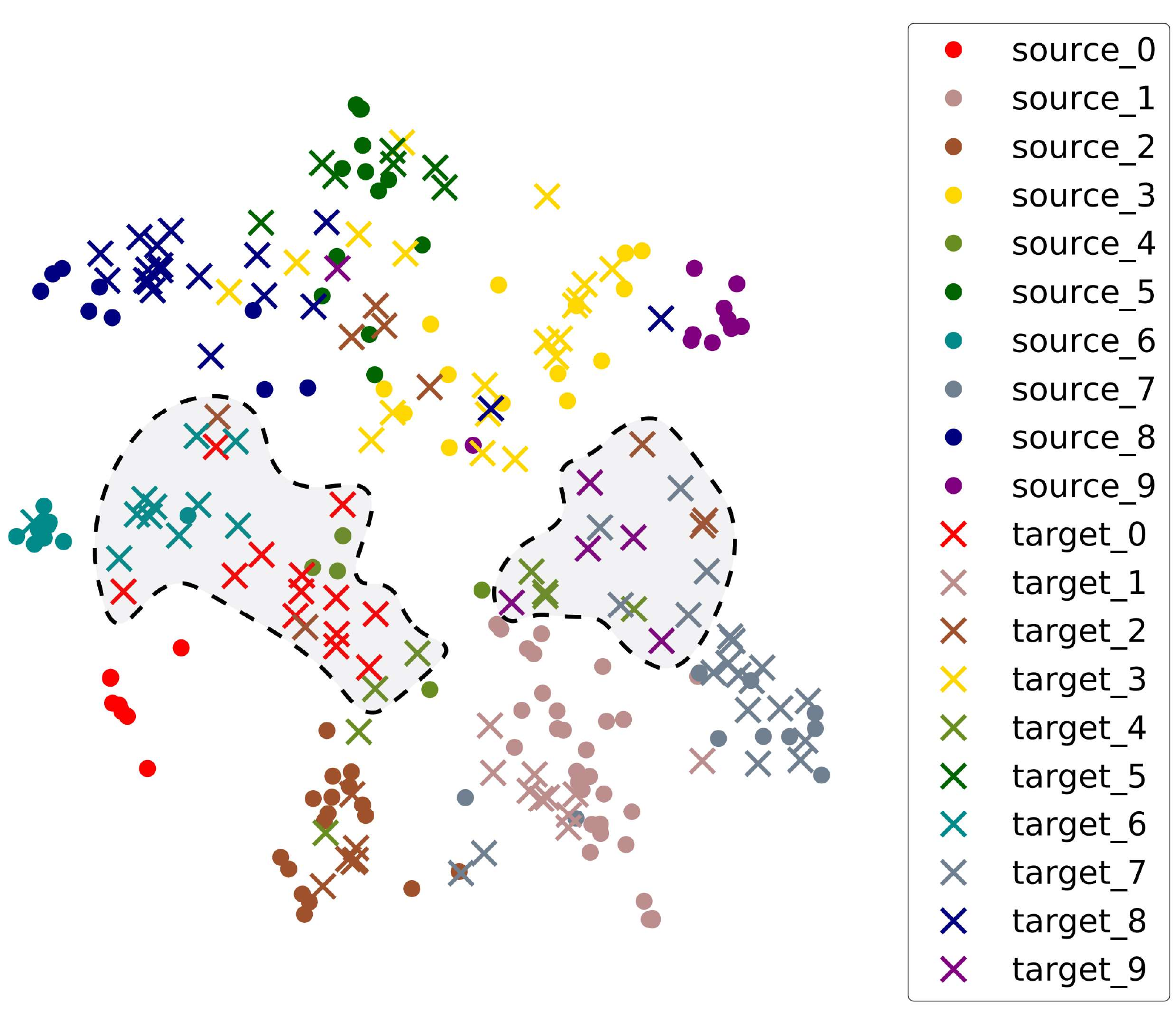} \\[\abovecaptionskip]
    \small (a) \textit{source only} model
  \end{tabular}

  \vspace{\floatsep}

  \begin{tabular}{@{}c@{}}
    \includegraphics[width=\linewidth,height=5cm]{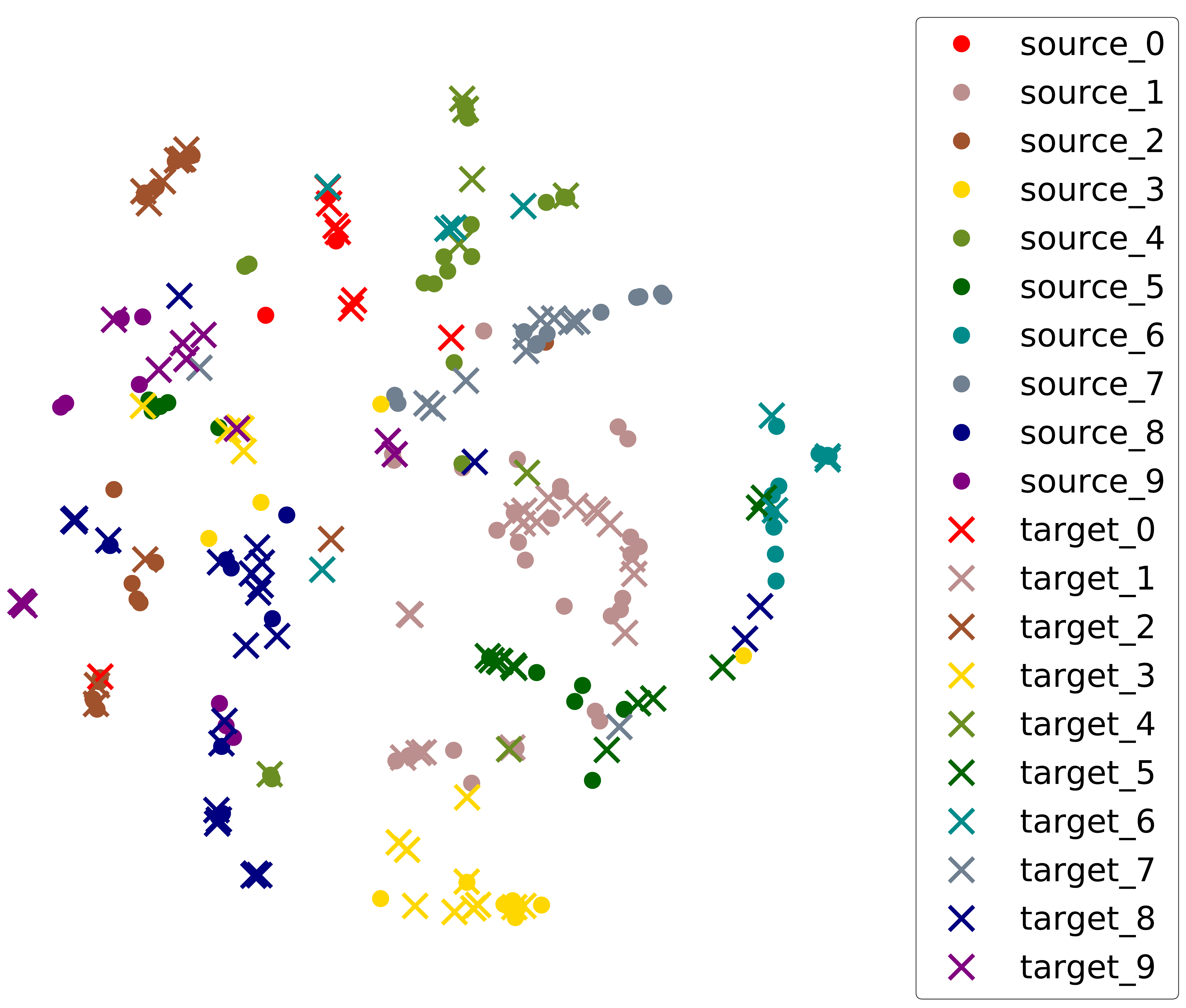} \\[\abovecaptionskip]
    \small (b) DANN model
  \end{tabular}

  \caption{\textbf{Visualization of features from the \textit{source only} model and DANN model in SVHN$\rightarrow$MNIST scenario.} Circle and x markers represent the source and target domain features, respectively. The label is indicated in the right legend. The target features in (a) relatively lie around the source features but are still distant especially for the labels 0,4,6, and 9 (the shadowed region). (b) shows more aligned features but they still do not fit exactly and the boundaries between classes are unclear.}\label{problem raise}
\end{figure}


Unsupervised domain adaptation has garnered particular interest for exploiting the capacity of machine learning beyond supervised learning. 
Although supervised learning has proven its high practicability in numerous fields such as speech recognition, object detection, and machine translation~\cite{machine_translation,speech_recognition,object_detection}, it requires massive amounts of labeled data. 
Unfortunately, many situations exist where collecting labeled data is unrealistic or costly. In the robotics field, obtaining labels may damage robots~\cite{pixelda}; in the manufacturing industry, inevitable and frequent process changes render it infeasible to obtain labeled data. 
Domain adaptation, in which another domain that can easily get the corresponding label and exhibits similar characteristics with the targeted data is used, was suggested to address the mentioned label issue.

Formally, domain adaptation aims to learn a classification task for one domain (called \textit{target domain}) using enough but unlabeled data, $\{x^t\}$, in the target domain and labeled data, $\{x^s,y^s\}$, from a similar but different domain (called \textit{source domain}).\footnote{We will denote the domain using superscripts and omit them when they are not necessary.}
Domain adaptation assumes that the similar domain data shares discriminative features that facilitates in learning a specific task~\cite{da_survey2}. 
However, exploiting only source data in a supervised manner to learn a task in a target domain~\cite{transferable} results in high accuracy in the source domain but low accuracy in the target domain~\cite{da_survey2} because of the difference between two domains' data distributions, called \textit{domain shift} (often called domain discrepancy).
\textit{Finetuning} on the target domain with a small set of labeled target data was suggested; however, it tends to overfit to small labeled data~\cite{da_survey}. 

To address the domain shift, researchers have proposed to obtain invariant features between two domains while still discriminative by explicitly reducing the distance in embedded feature distributions from the two domains. 
\cite{mmd_da} uses the Maximum Mean Discrepancy (MMD)~\cite{mmd} and \cite{dann} exploits Generative Adversarial Networks (GANs) \cite{gan} for this purpose and achieved excellent performance. 
Fig.~\ref{problem raise} shows the effect of reducing the domain shift.
Fig.~\ref{problem raise} (a) and Fig.~\ref{problem raise} (b) are the t-SNE results of the features from the supervised model using only the source data (called \textit{source only} model) and DANN model~\cite{dann}, respectively.
The features overlap more in Fig.~\ref{problem raise} (b) especially for the labels 0, 4, 6, and 9 (the shadowed region in Fig.~\ref{problem raise} (a)). 
It results in the significant accuracy increase from 57.1$\%$ for the source only model to 73.9$\%$ for the DANN model.


However, these methods can still be improved further. 
Because the model learns to extract the embedded feature, which is discriminative in the source domain, it results in clusters on the source features and the decision boundary is determined according to them.
Subsequently, the embedded features in the target domain should constitute similar clusters along with the source features to achieve high accuracy.
In other words, the target feature should gather around the source clusters so that the decision boundary does not cross the target features. 
However, the aforementioned methods, which reduce only the distance between two marginal distributions, let the features overlap only generally but not let each cluster satisfactorily match as shown in Fig.~\ref{problem raise} (b).
In addition, source features also do not form perfectly separable clusters.

We herein propose a novel domain adaptation method exploiting label propagation and cycle consistency; it enforces the clusters of the embedded features to overlap exactly and the source features to establish more divisible groups.
Concretely, we reinforce the correspondence between the original labels of the source domain and the propagated labels that are obtained by propagating labels from the source domain to the target domain and back to the source domain.  
The forced cycle consistency facilitates the clusters of features in two domains that are near to each other to be closer, and those that are far to be more diverged.
This is the desired property as the discriminative source features can be applied to the target domain.

We demonstrate empirically that the proposed method can address the domain shift between two datasets more clearly.
In addition, we demonstrate that our method results in embedded features from two domains to form exactly overlapped clusters in the t-SNE result, which serves our purpose.
As a result, our method achieves high performance in multiple datasets. 

\section{Related Work}
\textbf{Unsupervised Domain Adaptation}
\cite{bendavid} introduced the theoretical analysis of domain adaptation 
that the classification error on the target domain is bounded by that on the source domain, domain discrepancy, and difference in labeling functions. 
Based on this analysis, a number of works have endeavored to train domain-confusing features to minimize domain discrepancy \cite{deep_domain_confusion, jda, dan, dann, adda}.
\cite{deep_domain_confusion, dan} employed MMD as the measure of the domain discrepancy to achieve domain confusion.
Inspired by GANs,
~\cite{dann} 
converted a domain confusion task into a minimax optimization that trains a feature generator and binary domain classifier simultaneously and adversarially.

Although minimizing domain discrepancy might be effective to reduce the upper bound of the error, it does not guarantee that the feature representation of the target domain is sufficiently discriminative. 
Hence, \cite{dsn,mstn,similarity_da} proposed complementary loss. 
\cite{dsn} argued that separating shared representation and the individual characteristics of each domain explicitly could enhance the accuracy of the model. They thus proposed a network design with private/shared encoders and a shared decoder. 
\cite{mstn} and \cite{similarity_da} deployed the centroid and prototype of each category to attempt a class-level alignment.  
While the studies above focused on feature-space adaptation, approaches to directly convert target data to source data have also been introduced \cite{pixelda, cycada, dacil}.
Those proposed methods intend to transfer the style of images to another domain while preserving the content.
The image-oriented methods demonstrate excellent performances on datasets that are similar on the pixel level; however, they may be likely to fail when a mapping function between the high-level feature and the image is complex \cite{adda}.

\textbf{Metric Learning}
Metric learning is learning an appropriate metric to measure the similarity or distance between data \cite{metricsurvey}. 
To illustrate its benefits, if the distances between similar data are minimized and the distances between distinct data are maximized, the accuracy of a classifier can be strengthened~\cite{triplet}.

Metric learning is particularly beneficial when the amount of available labeled data is minute, as in the cases of semi-supervised learning and unsupervised learning. 
In unsupervised domain adaptation, \cite{song} combined metric learning and cyclic consistency.
They maximized the inner product of the source feature and target feature with same label whereas minimizes the similarity between features with different labels. 
\cite{ada} enforcesd the feature alignment between the source and target to bind them together by considering the transition from the source to the target, and vice versa, according to the learned similarity between features.
The alignment in feature space is achieved by forcing the round way transition probability to be uniform in the same class and to be minimal between different classes.
The stated methods above are effective in aligning the source and target domains; however, it seems that they rarely consider the relationship between the unlabeled samples.

Graph-based learning are closely related to metric learning in that it stimulates clustering using the distance information. 
It mostly assumes label consistency \cite{zhou} that adjacent data tends to have the same labels \cite{wang}. 
Label propagation \cite{zhou} has shown improved performance on semi-supervised learning by satisfying the label consistency through propagating labels from labeled data to unlabeled data. 
To overcome the limitation that graphs should be provided and fixed in advance, \cite{Label_propagation_network, label_propagation_fewshot} learned distances between each node adaptively as in the metric learning and demonstrated high accuracy in both semi-supervised learning and few-shot learning. 



\section{Method}
\begin{figure*}[t!]
\centering
\includegraphics[width=\linewidth, height=6.5cm]{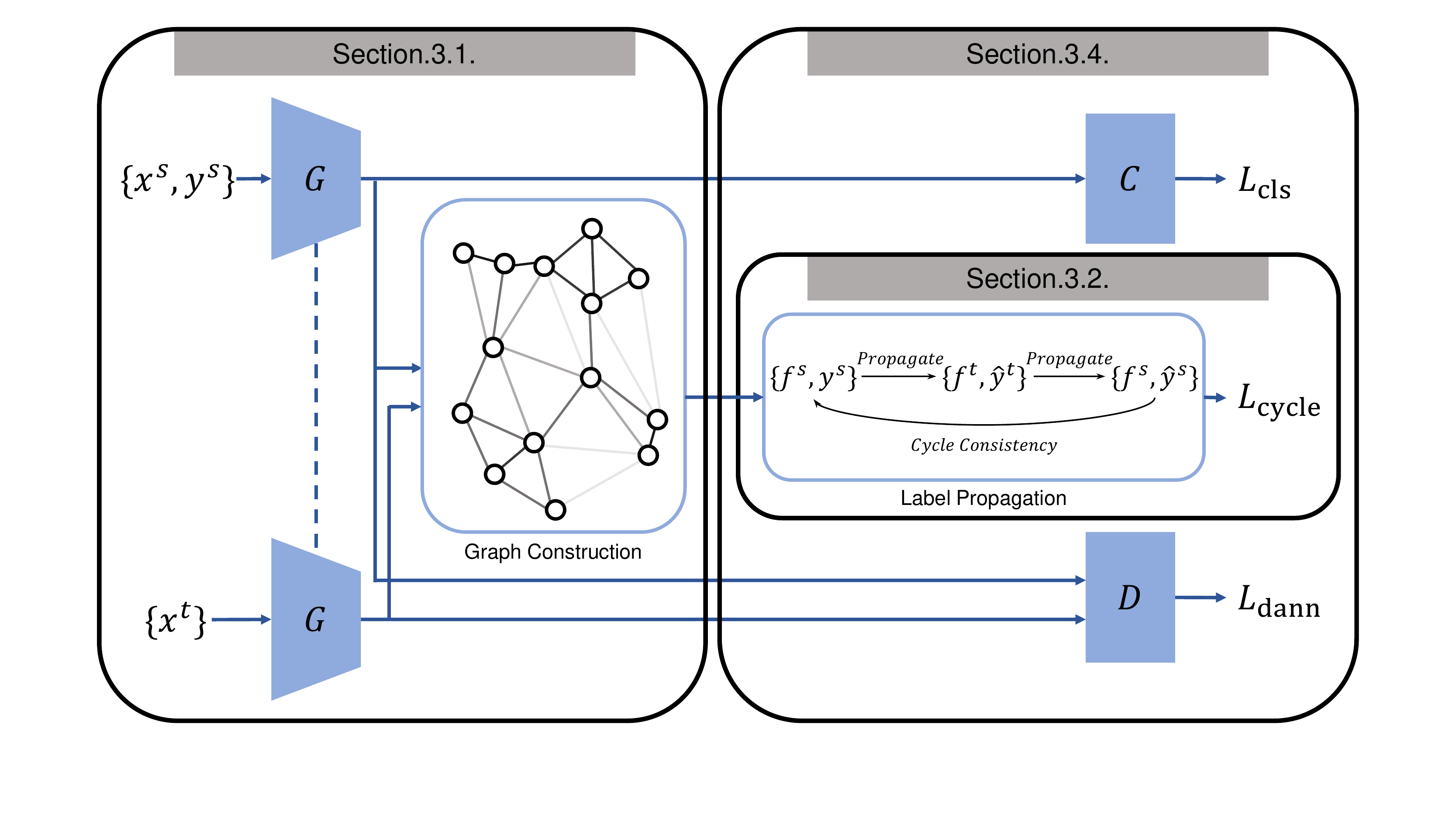}
\vspace*{-0.7cm}
\caption{\textbf{Overview of our proposed method.} The feature generator indicated by G projects the input data into the feature space; the graph is constructed using the embedded features. The graph is used to evaluate the cycle consistency with the label propagation yielding the cycle loss. The classifier (C) learns to classify the source features according to their ground-truth labels; the discriminator (D) takes the features from both domains and attempts to distinguish whether they come from the source or the target domain.}
\label{overview}
\end{figure*}

In this section, we introduce our novel algorithm to make embedded features from the two domains constitute the similar clusters.
Our method exploits label propagation and cycle consistency to learn embedded features $f(x^s)$ and $f(x^t)$ which are 1) indistinguishable from each other and 2) close within the same class and distant between different classes.
In addition, the model design of our method, illustrated in Fig.~\ref{overview}, contains the feature generator and the classifier which provides the final prediction follows it. 
The discriminator is used for DANN loss explained in Sec.~\ref{training process}.

\subsection{Feature Embedding and Graph Construction} \label{graph construction}
Manifold learning extracts low-dimensional embedded structures to learn a successful classifier. 
To obtain such structural information, it is typical to use a graph of which edges indicate the relation between each data. 
In our method, we build a graph using the embedded features. 
We first embed the input to the feature space by employing neural networks with a few convolutional layers, as in \cite{Label_propagation_network, label_propagation_fewshot} (Feature generator in Fig.~\ref{overview}). 
Subsequently, we construct a fully connected graph between the feature based on their distance to each other. 
For the similarity weight for each edge, we used the typical choice, \ie, Gaussian similarity. 
The edge weight between the input data $x_i,x_j$ is expressed by the following similarity weight:
\begin{align}
W_{ij} = \exp(-\frac{\left\lVert f_i-f_j \right\rVert^2}{2\sigma^2})\label{similarity weight}
\end{align}
where $f_i,f_j$ are the embedded feature vectors of $x_i,x_j$. 
The Euclidean distance in Eq.~\ref{similarity weight} can be replaced by any distance such as the L1 distance or cosine similarity; however, we empirically found that the performance of the Euclidean distance is superior to those of others.
$\sigma$ is the scale parameter which is known that graph-based methods are sensitive to \cite{label_propagation_fewshot}.
A large $\sigma$ results in a uniformly connected graph that disregards the latent structure, while a small $\sigma$ renders the graph sparse and hinders the relation between data to be obtained. 
We thus train $\sigma$, which has the same dimensionality as a feature, adaptively.

\subsection{Label Propagation and Cycle Consistency}
Label propagation is suggested to implement manifold regularization; it constraints the classifier to be smooth so that it should not change its prediction significantly for a small perturbation.
Label propagation, central part of our method, can be interpreted as repeating random walk using the similarity matrix infinitely to assign the labels of target data.

Label vector $y_n\in \mathbb{R}^{(\mathrm{N}_s+\mathrm{N}_t)\times \mathrm{C}}$ indicates assigned labels for both domain data at the n-step random walk. 
$\mathrm{N}_s$,$\mathrm{N}_t$, and $\mathrm{C}$ represent the number of source data, target data, and classes, respectively.
The first $\mathrm{N}_s$ rows, $y_n^s$, refer to the labels of the source data, and the remaining rows, $y_n^t$, present the labels of the target data (\ie, $y_n = [y_n^s;y_n^t]$). 
The label vector $y_n$ is initialized to one-hot coded ground-truth labels for the source data and a zero vector for the target data. 
The one-step random walk transforms the label vector as follows:
\begin{align}
    y_{n+1} &= T{y_n}\\
    \mathrm{where,}\: T &= 
    \begin{pmatrix}
    I & 0 \\
    T_{ts} & T_{tt}
    \end{pmatrix} = normalize(W),\\
    W &= \begin{pmatrix} 
I & 0 \\
W_{ts} & W_{tt}
\end{pmatrix}.
\end{align}
The $normalize(\cdot)$ is a normalizing operation that transforms the sum of each row to 1.
$W_{ts}$ is the similarity weight matrix between the target data and source data as described in Sec.~\ref{graph construction} and $W_{tt}$ is the similarity weight matrix between the target data and themselves. 
$I$ and $0$ are the identity and zero matrices, respectively. 
$T$, the normalized $W$, is the transition matrix in which $I$ means that the source data do not move (\ie, source data nodes are absorbing nodes from the graph theory perspective), as the label for the source data is already known and fixed. 
The label propagation for the target domain data is referred to the propagating label by infinite transition as follows:

\begin{align}
    {\hat{y}}^t &= \lim_{n->\infty}\mathrm{Last\:N_t\:rows\:of}\:T^n y_0 \label{lp} \\
      &= T_{tt}y_0^s+ T_{tt}T_{ts}y_0^s+T_{tt}T_{tt}T_{ts}y_0^s + \cdot\cdot\cdot \label{unfold lp} \\
      &= (I-T_{tt})^{-1}T_{ts}y_0^s. \label{closed form lp}
\end{align}

As the source data does not transition to the other data, provided that $W_{ts}$ is not 0, Eq.~\ref{lp} converges to a closed form, Eq.~\ref{closed form lp}. 
The obtained label vector ${\hat{y}}^t$ is used to predict the label of the target data. 
In our method, it is used to learn the features of which their clusters match each other.

As we obtain the label of the target data ${\hat{y}}^t$ propagated from the source data using Eq.~\ref{closed form lp}, we can attain the propagated label of the source data ${\hat{y}}^s$ in turn, using ${\hat{y}}^t$ as follows:
\begin{align}
    {\hat{y}}^s = (I - T_{ss})^{-1}T_{st}{\hat{y}}^t.
\end{align}

Now we denote the cycle consistency by the desired property that ${\hat{y}}^s$ should be the same as the original label $y^s$. 
If the embedded features from both domains form well-aligned clusters, cycle consistency should be realized; otherwise, it may not. 
We thus speculate that by achieving the property, the performance may improve. 
To establish cycle consistency, the model was enforced to minimize the following L1 cycle loss between ${\hat{y}}^s$ and $y^s$:
\begin{align}
    L_\mathrm{cycle} = \left\lVert {\hat{y}}^s - y^s \right\rVert_1 \label{cycle loss}
\end{align}

\begin{figure}[t]
\begin{center}
\centerline{\includegraphics[width=\columnwidth,height=\columnwidth]{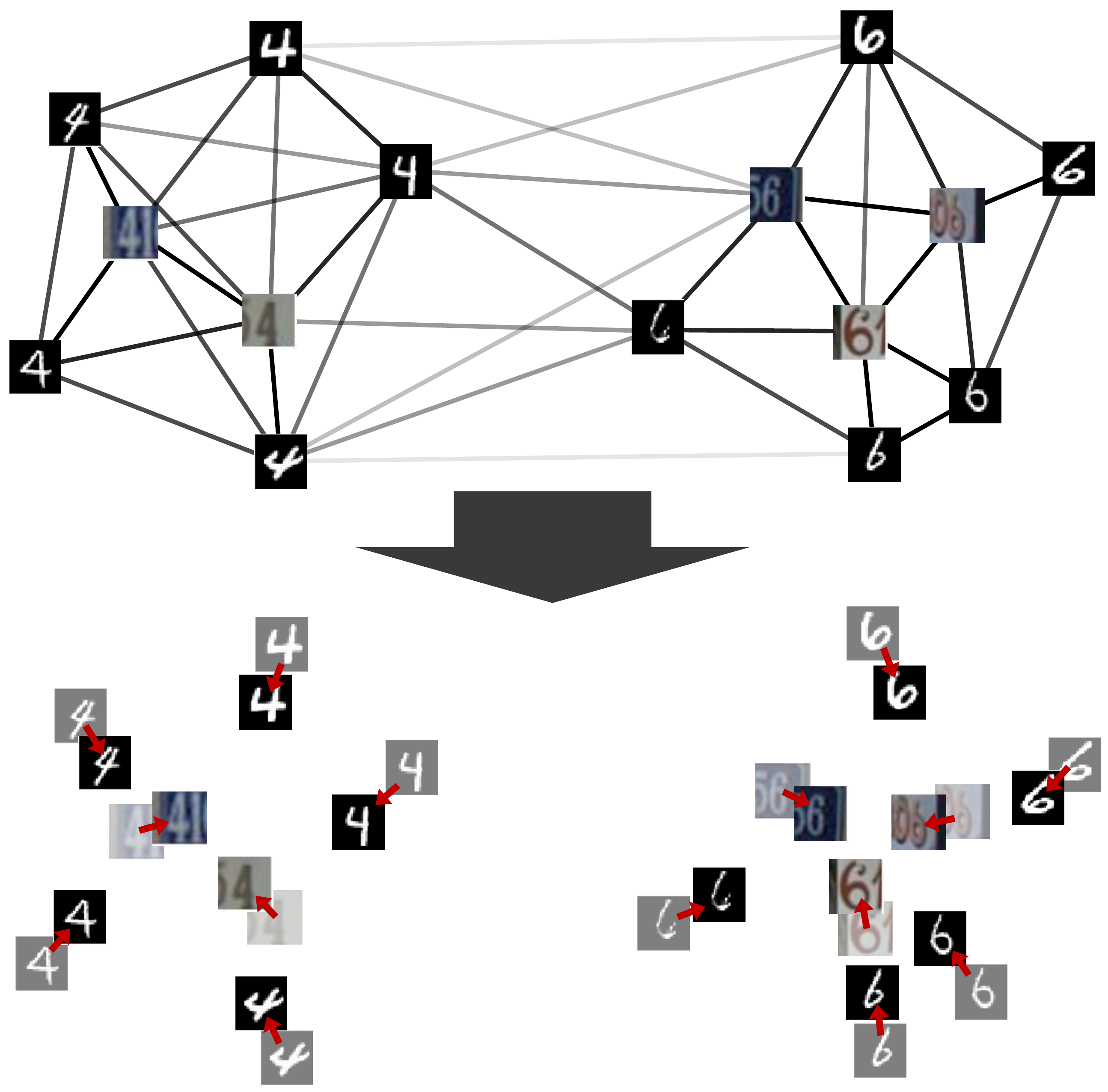}}
\caption{\textbf{Graphical description of the effect of cycle loss.} An example of the cycle loss effect is depicted for SVHN$\rightarrow$MNIST scenario. The upper figure illustrates the constructed graph, and the darkness of lines refer to the similarity between two samples. Strong relationship between a source sample and a target sample leads $\hat{y}^t$ to be aligned with $y^s$, and then $\hat{y}^s$ to be aligned with $\hat{y}^t$ for the same reason. The relationship is determined not only by direct connection but also indirect path along other nodes. To attain cycle consistency, each feature receives power to gather within the same class and to get far away from the other classes. As a result, feature representation becomes class-wisely clustered as portrayed in the lower figure.}
\label{moving_illustration}
\end{center}
\vskip -0.8cm
\end{figure}

Minimizing $L_\mathrm{cycle}$ may result in a high performance because not perfectly aligned features are forced to move toward a near cluster, as illustrated in Fig.~\ref{moving_illustration}.
Cycle consistency is the typically used concept to match one to another. 
For example, \cite{cycada} compares pixel-wise values between a cyclic-generated image and the original input image, but it merely matches one to another, and does not exploit the entire manifold structure.
With the structure information, the embedded features are forced to move to the clusters, instead of to one point.

\subsection{Controlling the Weight of Gradient} \label{gradient weight}
The objective of our method is to make embedded features from two domain constitute overlapped clusters.
To pursue this objective, each feature is categorized into three cases for analysis;
1) Has no dominant closest cluster,
2) Moderately close to a cluster,
3) Perfectly absorbed to a cluster.
When the model has not learned the discriminative features sufficiently, some data may be confusing among two or more classes, and their features would be embedded in somewhere between the clusters. 
In this case, drawing these features to any closest but still distant clusters may cause noise in learning process. 
In contrast, the model may have nothing to learn from drawing already well-aligned features to their assigned clusters. 
We thus suggest that it is important to align features that lie around the clusters but not fit perfectly to them.
In this regard, weight for the gradient with respect to each feature is specified according to its case.

We allowed for the effect of each data on the learning process through the cycle loss to be different by multiplying the weight to the derivative of the cycle loss with respect to the embedded features, thereby resulting in the gradient for trainable variable $w$ as following modified chain rule:
\begin{align}
    \frac{dL_\mathrm{cycle}}{dw} &= \sum_i \rho_i \frac{dL_\mathrm{cycle}}{df_i}\frac{df_i}{dw} \label{grad weight} \\
    \mathrm{where,} \: \rho_i &= \mathsf{e}H(p_i)\exp(-H(p_i)). \label{mountain shape}
\end{align}
where $f_i$ is the embedded feature and $\rho_i$ is the weight for the $i$-th data. $p_i\in\mathbb{R}^{\mathrm{C}}$ is the predicted probability for the $i$-th data and $H(\cdot)$ is the entropy of given probability distribution. 
Euler's number $\mathsf{e}$ is multiplied to adjust the scale. 
The bell-shaped $\rho_i$ with respect to $H(p_i)$ in Eq.~\ref{mountain shape} is defined such that features that 1) has no dominant closest cluster or are 3) perfectly absorbed to a cluster may not affect the learning process significantly.

\subsection{Training Process} \label{training process}
We need to learn features discriminative for the task from the labeled source data and indiscriminative between two domains according to the analysis of \cite{bendavid}. 
The overall training loss for our model is as follows:
\begin{align}
    L = L_\mathrm{cls} + L_\mathrm{dann} + \alpha L_\mathrm{cycle}. \label{loss}
\end{align}
$L_\mathrm{cls}$ is defined as the typically used cross entropy loss using labeled source data and $L_\mathrm{dann}$ is defined as the original GAN loss~\cite{gan}.
\begin{align}
    &L_\mathrm{cls} = \frac{1}{\mathrm{N}_s}\sum_{i=1}^{\mathrm{N}_s}-y_i^s \log{p_i(y=y_i^s)}.\\
    & {L_\mathrm{dann}} = \frac{1}{\mathrm{N}_s}\sum_{i=1}^{\mathrm{N}_s}\log{D(f_i^s)} + \frac{1}{\mathrm{N}_t}\sum_{j=1}^{\mathrm{N}_t}\log{(1-D(f_j^t))}
\end{align}
where $D(\cdot)$ is the output of the discriminator with value in the range of [0,1]. All components but the discriminator are trained to reduce the overall training loss, and the discriminator learns to maximize $L_\mathrm{dann}$.

From the metric learning perspective, $L_\mathrm{cls}$ serves to separate the source features according to their ground-truth labels. As shown in Fig.~\ref{source only embedding}, the source features are likely to congregate according to their classes. 
Subsequently, $L_\mathrm{dann}$ takes a role to move the target features toward the source features, but it is insufficient to lead the perfectly aligned clusters. 
Our cycle loss $L_\mathrm{cycle}$ facilitates in realizing it by enforcing the cycle consistency. 
The experimental results on several domain adaptation scenarios and visualized embedding results are described in the next section to support our argument.

\section{Experiments}

\begin{table*}[h]
\small
  \centering
    \caption{\textbf{Accuracy(\%) evaluation of digit domain adaptation tasks.} Most results reported below are extracted from \cite{pixelda} and \cite{adda}. For the DANN results on MNIST$\rightarrow$USPS and USPS$\rightarrow$MNIST, which are not reported in the original paper, the evaluation conducted by \cite{adda} are shown. $^\ast$ represents the result obtained by executing the publicized code.}    
    \begin{tabular}{r|@{\hskip 0.7in}c@{\hskip 0.7in}c@{\hskip 0.7in}c@{\hskip 0.7in}c}
\toprule
      \small{Source} & \small{MNIST} & \small{MNIST} & \small{USPS} & \small{SVHN} \\
      \small{Target} & \small{MNIST-M} & \small{USPS} & \small{MNIST} & \small{MNIST} \\

\midrule
      \small{Source Only} & 63.6 & 75.2 & 57.1 & 60.1 \\ 
      \small{MMD~\cite{mmd_da}} & 76.9 & 81.1 & - & 71.1 \\ 
      \small{DANN~\cite{dann}} & 76.7 & 77.1 & 73.0 & 73.9 \\ 
      \small{DRCN~\cite{drcn}} & - & 91.8 & 73.7 & 82.0 \\   
      \small{CoGAN~\cite{cogan}} & 62.0 & 91.2 & 89.1 & - \\ 
      \small{ADDA~\cite{adda}} & - & 89.4 & 90.1 & 76.0 \\
      \small{DSN w/ MMD~\cite{dsn}} & 80.5 & - & - & 72.2 \\
      \small{DSN w/ DANN~\cite{dsn}} & 83.2 & - & - & 82.7 \\
      \small{kNN-Ad~\cite{song}} & 86.7 & - & - & 78.8 \\
      \small{AssocDA~\cite{ada}} & 89.5 & - & - & 97.6 \\
      \small{PixelDA~\cite{pixelda}} & 98.2 & 95.9 & 97.8$^\ast$ & - \\
      \small{Cycada~\cite{cycada}} & - & 95.6 & 96.5 & 90.4 \\
      \small{ATT~\cite{att}} & 94.2 & - & - & 86.2 \\
      \small{LEL~\cite{lel}} & - & - & - & 81.0 \\
      \small{SimNet~\cite{similarity_da}} & 90.5 & 96.4 & 95.6 & - \\
      \small{MSTN~\cite{mstn}} & - & 92.9 & - & 91.7 \\
      \midrule
      Ours & 96.2 & 95.7 & 98.8 & 94.3 \\
      \bottomrule
    \end{tabular}%
  \label{visual_domain_result}%
\end{table*}

In this section, we evaluate our algorithm on the visual digit datasets and Amazon review datasets. 
We first describe the experimental setup and then present our results. 
In addition, we support the validity of our method by presenting the t-SNE result of the embedded features.

\subsection{Implementation Details}

\textbf{Architecture} 
A neural network architecture with two convolutional layers and two fully connected layers as in \cite{mstn} was used for all experiments.
All the experiments were implemented using Tensorflow \cite{tensorflow}. 
The Gradient Descent optimizer with a momentum of 0.9 was utilized for the training \cite{momentum}.

\textbf{Learning Rate}
A learning rate of $10^{-2}$ was chosen for all experiments with the support of the batch normalization technique that stabilizes gradient and thus enables a higher learning rate \cite{batchnormeffect}.

\textbf{Delay of $L_{cycle}$}
To reduce the adverse effect of noisy gradients from the cycle loss at the early stages of training, a weight balance parameter, $\lambda = \frac{2}{1+exp(-\gamma \cdot p)}-1$ is multiplied to the cycle loss in line with Sec.~\ref{gradient weight}.
$\gamma$ refers to a velocity constant that determines the rate of increase of $\lambda$; p is the training progress that changes from 0 to 1.
The parameter was first introduced in \cite{dann} such that the classifier is less sensitive to the erroneous signals from the discriminator in the beginning.
Throughout the experiments, $\gamma$ was set to 10.

\textbf{Hyperparameter}
Although it would be ideal to avoid utilizing labels from the target domain in the hyperparameter optimization, it seems that no globally applicable method exits for this.
\cite{dann} proposed the reverse validation scheme, but \cite{dsn} found that the reverse validation accuracy often does not match the test accuracy.
In addition, as \cite{dsn} stated, applications exist where the labeled target domain data are available at the test phase but not at the training phase.
Hence, by taking this observation, a small set of labeled target domain data was exploited as a validation set: 1152 samples for visual domain adaptation and 256 samples for the Amazon review experiment, similar to \cite{att,dsn,pixelda}.

\textbf{Batch Sizes}
Owing to the inherent characteristics of label propagation that each sample data affects the graph structure, it is important for each class sample in each batch to represent its classes properly.
In other words, the transition matrix might be erroneous if biases exist in the samples.
Therefore, the data samples per each class in a batch should be sufficient to circumvent the probable biases.
To address this problem, we performed experiments with batch sizes of up to 384 and observed minute improvement beyond batch size of 128.
In this context, all of the domain adaptation experiments were conducted with batch size of 128.

\subsection{Adaptation Settings}

\textbf{MNIST} $\rightarrow$ \textbf{MNIST-M} 
MNIST is the hand-written digit images of 10 classes \cite{mnist}, and MNIST-M is a variation of MNIST proposed by \cite{dann}. Specifically, MNIST-M was created by blending MNIST to random crops from the BSDS500 dataset \cite{bsds}.
In addition, similar to \cite{similarity_da}'s settings, MNIST images were inverted randomly for this scenario, because their colors are always white on black, whereas the MNIST-M images exhibit various colors.

\textbf{MNIST} $\leftrightarrow$ \textbf{USPS} 
USPS \cite{usps} is another hand-written digit image dataset with 10 classes and can be used for domain adaptation with MNIST. 
USPS contains 16$\times$16 images and the size of the USPS image is upscaled to 28$\times$28, which is the size of the MNIST image in our experiment.

\textbf{SVHN} $\rightarrow$ \textbf{MNIST} 
The discrepancy between MNIST and Street View House Numbers \cite{svhn} is greater than those of the previous scenarios because MNIST is a hand-written image dataset while SVHN is a real picture image dataset. 
The size of each MNIST image is upscaled to 32$\times$32, which is the size of SVHN images.

\textbf{Amazon Reviews}
To investigate the effectiveness of the proposed method on a non visual domain adaptation setting, the Amazon reviews dataset was employed. 
This dataset was created and processed by \cite{amazon} for sentiment domain adaptation experiment. Specifically, reviews regarding books, dvds, electronics, and kitchens are encoded in 5,000-dimensional feature vectors that are unigrams and bigrams with binary labels. 
If the review is ranked with four or five stars, a positive label is attached to it.
In contrary, if the review is ranked up to three stars, a negative label is attached to it.
For the training session, 2,000 labeled source data and 2,000 unlabeled target data were used.
For the testing, between 3,000 to 6,000 target samples were used.

\begin{table}[t]
\small
  \centering
    \caption{\textbf{Classification accuracy(\%) on Amazon Reviews experiments.} The accuracy of VFAE\cite{vfae}, DANN\cite{dann}, and ATT\cite{att} are shown with our result for comparison.}
    \begin{tabular}{cc|ccc|c}
    \toprule
    Source & Target & VFAE & DANN & ATT & \small{Ours}\\

    \midrule
    \small{books} & \small{dvd} & 79.9 & 78.4 & 80.7 & {81.3}\\[1.3pt]
    \small{books} & \small{electronics} & 79.2 & 73.3 & {79.8} & 78.3\\[1.3pt]
    \small{books} & \small{kitchen} & 81.6 & 77.9 & {82.5} & 79.7\\[1.3pt]
    \small{dvd} & \small{books} & 75.5 & 72.3 & 73.2 & {77.2}\\[1.3pt]
    \small{dvd} & \small{electronics} & 78.6 & 75.4 & 77.0 & {79.0}\\[1.3pt]
    \small{dvd} & \small{kitchen} & 82.2 & 78.3 & {82.5} & {82.5}\\[1.3pt]
    \small{electronics} & \small{books} & 72.7 & 71.1 & {73.2} & 70.8\\[1.3pt]
    \small{electronics} & \small{dvd} & {76.5} & 73.8 & 72.9 & 73.3\\[1.3pt]
    \small{electronics} & \small{kitchen} & 85.0 & 85.4 & 86.9 & {87.1}\\[1.3pt]
    \small{kitchen} & \small{books} & 72.0 & 70.9 & {72.5} & 71.8\\[1.3pt]
    \small{kitchen} & \small{dvd} & 73.3 & 74.0 & {74.9} & 73.5\\[1.3pt]
    \small{kitchen} & \small{electronics} & 83.8 & 84.3 & 84.6 & {85.4}\\[1.3pt]
    \midrule
    \end{tabular}%
  \label{amazon_result}%
\end{table}

\subsection{Results}

\textbf{Visual Domain Adaptation}
Tab.~\ref{visual_domain_result} compares the accuracy of our method on the visual digit adaptation experiment with other approaches.
On the MNIST$\rightarrow$MNIST-M task, the proposed algorithm demonstrates better accuracy compared to the others, except PixelDA.
PixelDA learns conversion in a pixel level from one domain to another, similar to how the MNIST-M dataset is generated.
Hence, it is likely that PixelDA shows extraordinary performance on the task because of its methodology. In addition, Cycada and PixelDA require a much larger network for image translation. Our method achieved a better or comparable accuracy with a much smaller architecture.
In the USPS$\rightarrow$MNIST scenario, our method outperforms all of the other methods. 
Specifically, we could observe that our method achieves a 25.8\% margin of improvement compared to DANN. 
This may imply that enforcing clustering in addition to domain-invariant embedding was essential in reducing the error rate.
The performance of our method is better or comparable than those of others in the MNIST$\rightarrow$USPS and SVHN$\rightarrow$MNIST experiments in general. 
Overall, we argue that our proposed method is outstanding and more robust compared to the other methods. 

\textbf{Adaptation in Amazon Reviews}
The results in Tab.~\ref{amazon_result} shows that the proposed method performs better than DANN~\cite{dann}, VFAE~\cite{vfae} and ATT~\cite{att} in six out of twelve experiments.
In particular, compared to the accuracy of DANN, the proposed method demonstrates better accuracy in nine out of twelve settings, and outperforms it by approximately 2.0\% on average.
From the result, we may argue that our intention to expedite the discriminative feature embeddings of the target domain was effective in reducing classification error in the target domain.


\section{Discussion}
\begin{figure*}
\centering
\begin{tabular}{cc}
\includegraphics[width=.8\columnwidth,height=4cm]{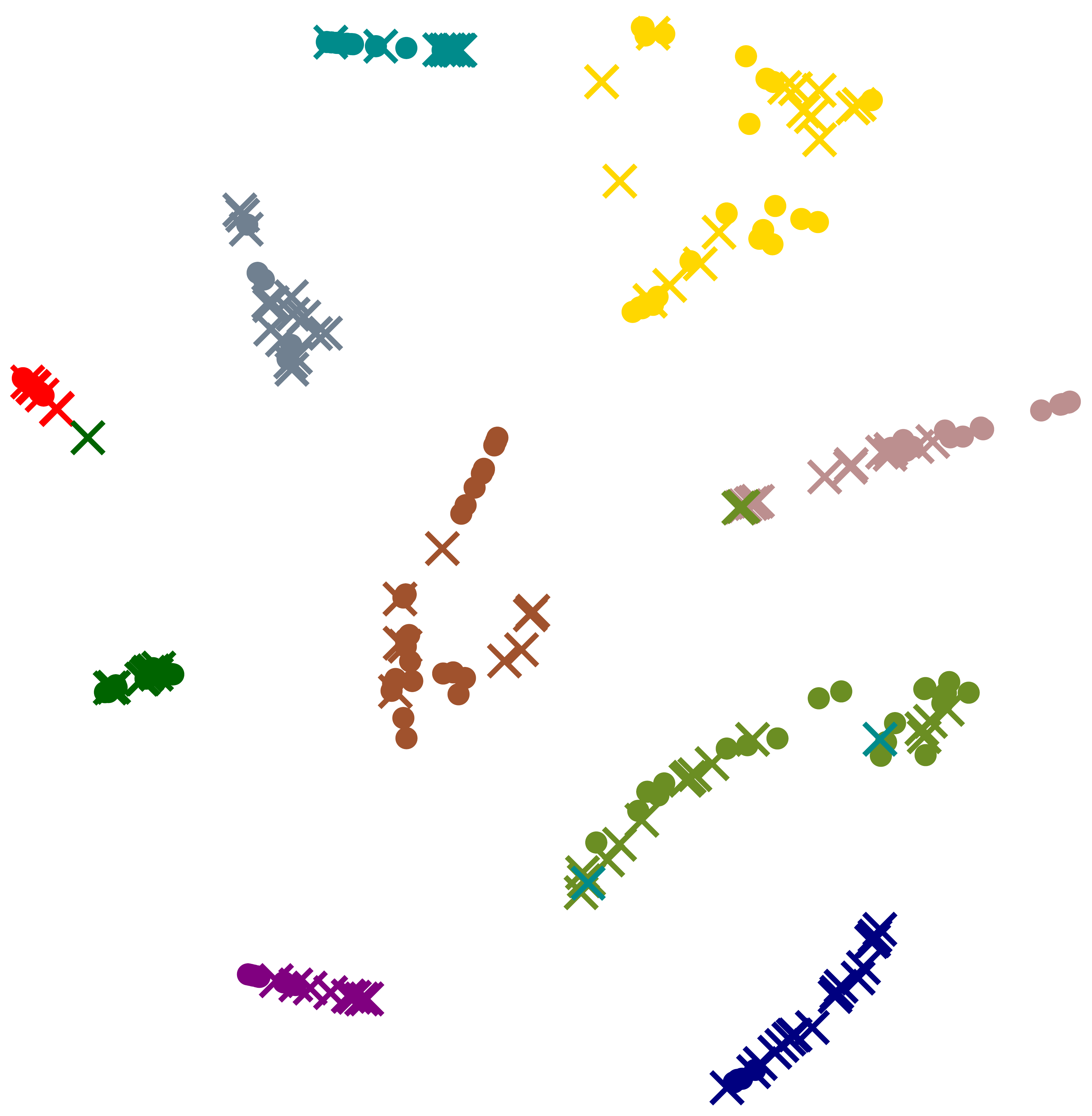} & \includegraphics[width=.8\columnwidth,height=4cm]{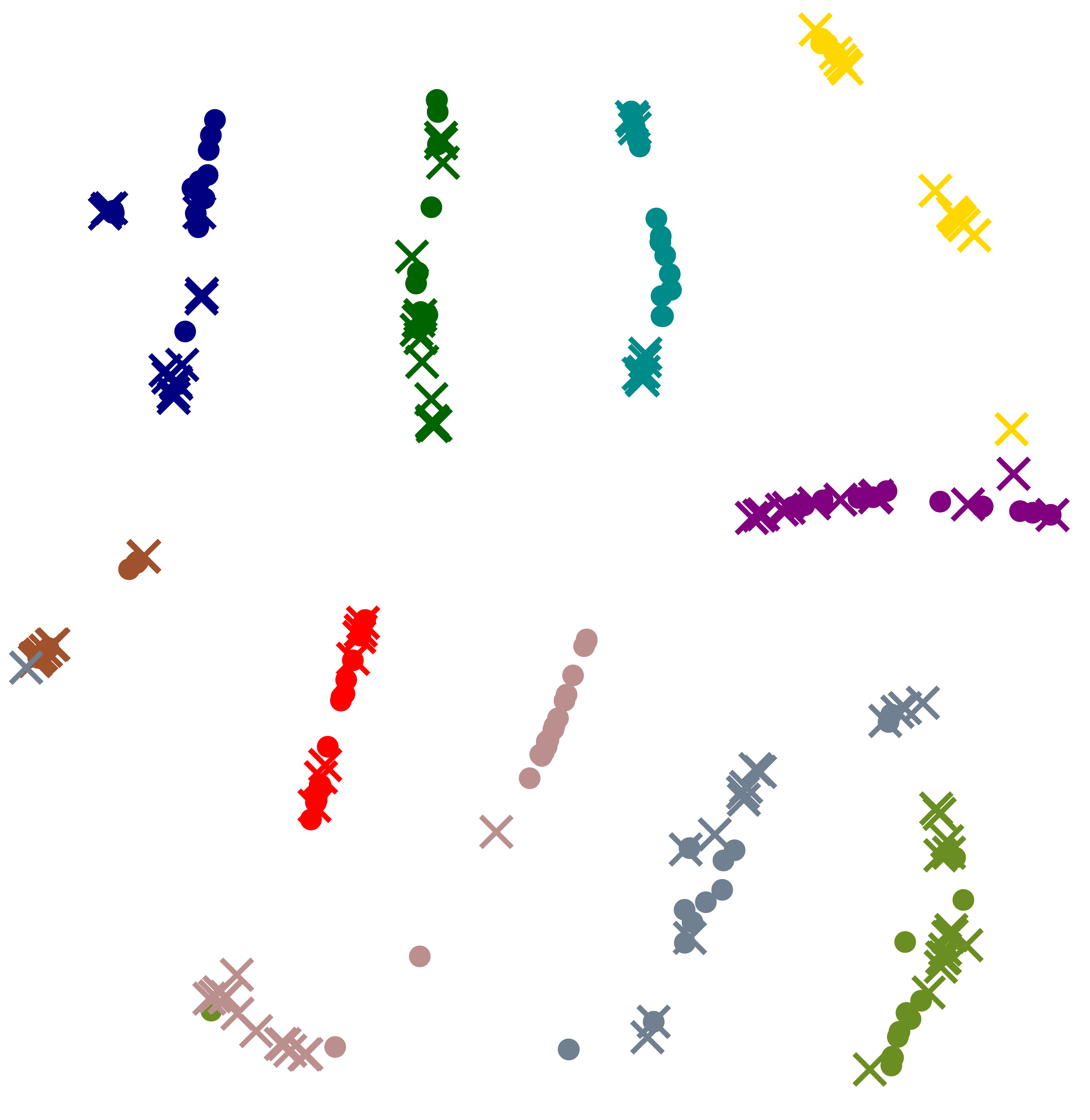}\\
\small (a) SVHN $\rightarrow$ MNIST & \small (b) USPS $\rightarrow$ MNIST\\ 
\includegraphics[width=.8\columnwidth,height=4cm]{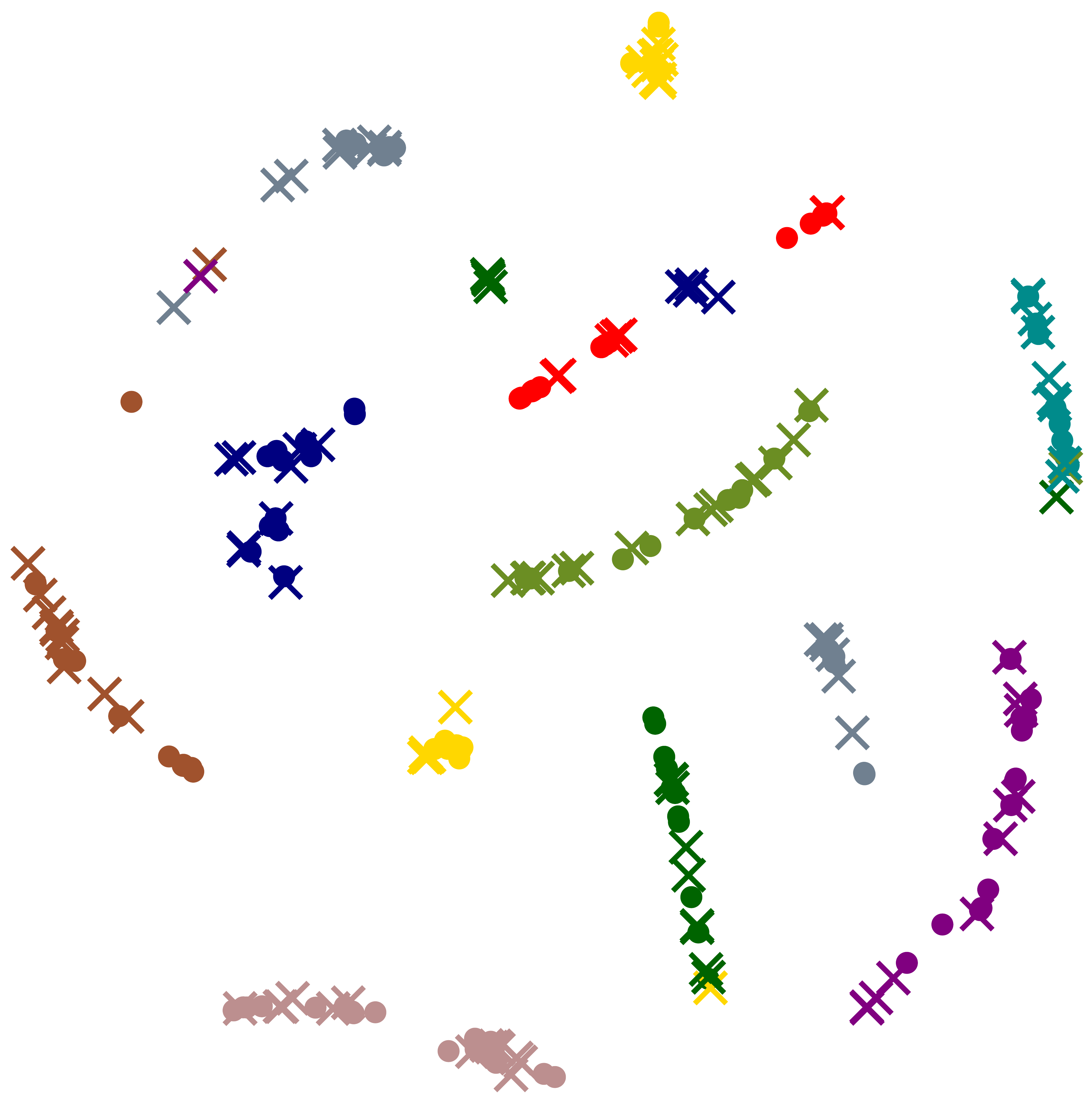} & \includegraphics[width=.8\columnwidth,height=4cm]{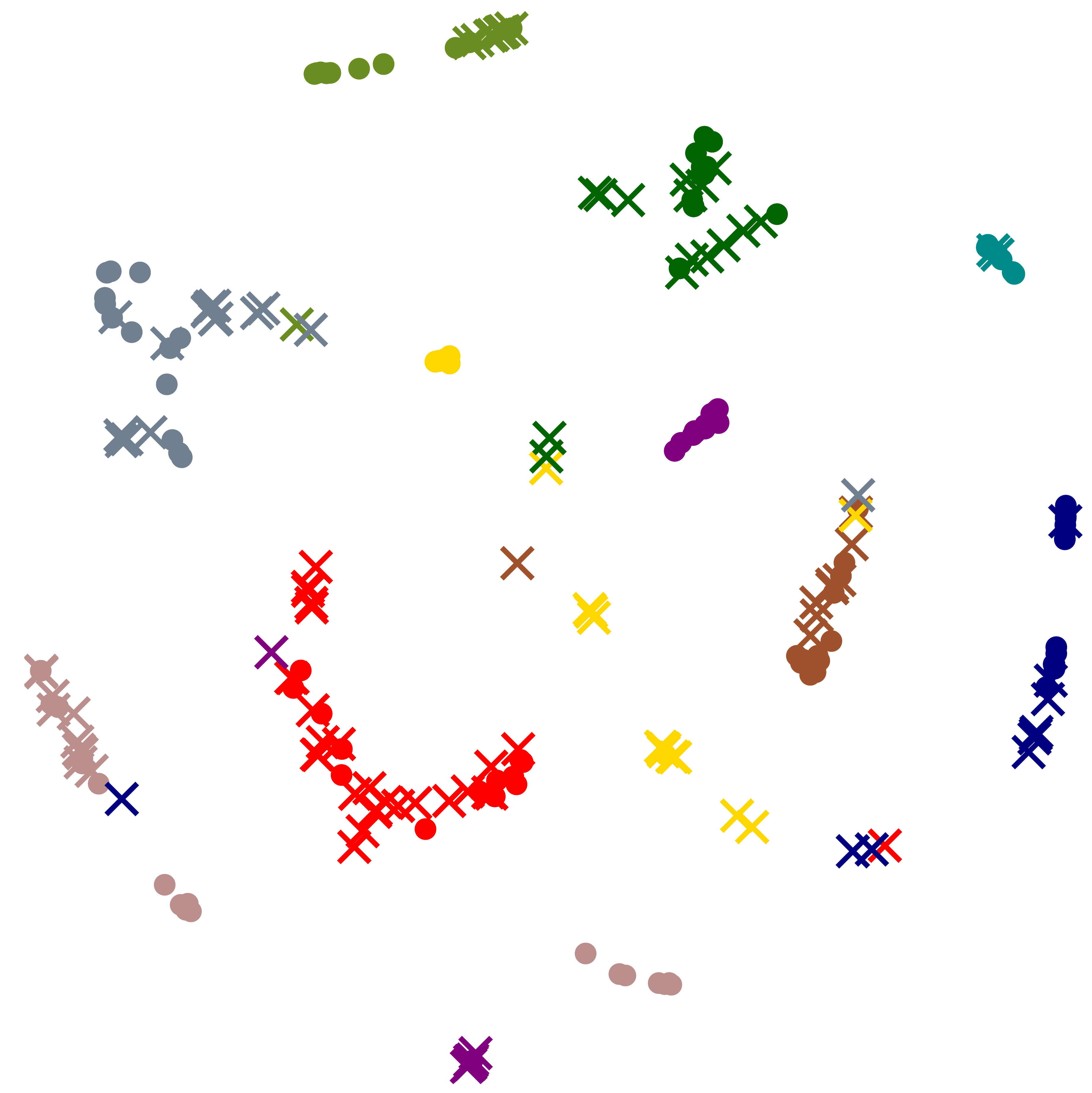} \\
\small (c) MNIST $\rightarrow$ MNIST-M & \small (d) MNIST $\rightarrow$ USPS \\
\end{tabular}
\caption{\textbf{Visualization of features from our method.} Circle and x markers indicate source and target features, respectively and colors indicate the labels. In all cases, the features overlap between two domains and exhibit clear clusters.}
\label{ourmethod}
\end{figure*}

We herein propose the novel domain adaptation using label propagation and cycle consistency. 
Our objective is to allow for the embedded features from the two domains to establish overlapped and condensed clusters in the feature space, in addition to staying near the clusters which can be obtained by reducing domain shift as \cite{dann,mmd_da}. 
The proposed method achieves our objective, and results in better domain adaptation for various scenarios compared to the existing methods. 
The t-SNE result in Fig.~\ref{ourmethod} proves that our method is critical in accomplishing our intended goal.


Cycle consistency is typically used in various domain adaptation methods to match the relation between the data and reinforce their adhesion \cite{cycada, song, ada}.
Our method also uses cycle consistency for the same purpose. 
However, unlike other methods, our method considers a manifold structure to evaluate and increase the cycle consistency using label propagation that is suggested as a manifold regularization in semi-supervised learning.
Apart from the direct relationship between each data, our method considers the indirect effect of the manifold; hence, the connection between the data in the same domain changes to satisfy the cycle consistency.
This induces not only congregated features between the two domains, but also condensed source feature clusters, which can be observed by comparing Fig.~\ref{problem raise}, where source features scatter relatively with Fig.~\ref{ourmethod}, where the source features can be separated perfectly.

While exploiting label propagation to obtain manifold information, matrix inversion is required for the implementation, as described in Eq.~\ref{closed form lp}.
One may doubt that the massive computation load for the matrix inversion, which is $O(n^3)$ for an $n\times n$ matrix, may degrade the efficacy of our method.
However, it was suggested that matrix inversion can be achieved with as the same cost as the matrix multiplication.
A sophisticated algorithm can reduce operation cost to $O(n^{2.373})$ ~\cite{inversion_complexity}, and it can be processed in parallel \cite{parallel_inversion} to enable a faster implementation. 
In addition, the numbers of rows and columns of the matrix to be inverted in Eq.~\ref{closed form lp} are equivalent to the batch size.
Considering the typically used batch size (64,128,256), it is affordable to perform the matrix inverse as the elapsed time for inverting those size of a matrix is up to approximately 0.2s.


To render a larger batch size feasible, we can approximate the label propagation using Eq.~\ref{unfold lp}. 
It is known that the random walk process converges in 10$\sim$20 steps~\cite{label_propagation_fewshot}.
Thus, by using the first a few terms in Eq.~\ref{unfold lp}, we can approximate the label propagation; further, it is still learnable through gradient-based optimization with slight performance degradation~\cite{label_propagation_fewshot}. 
In particular, caution should be exercised in applying the approximation to our method.
As the label $y^s$ in the cycle loss in Eq.~\ref{cycle loss} is one-hot coded, the propagated label $\hat{y}^s$ are likely to become all zeros, because it yields a lower loss than assigning the wrong labels. 
Consequently, it prevents propagating labels from one domain to another domain and results in the distant features between each domain, which is not desired. 
Therefore, we should constraint the sum of $\hat{y}^s$ to be one to penalize not assigned labels.

Our work focuses on embedding features to be discriminative between each group and indistinguishable between two domains, to achieve the better domain adaptation performance. 
However, our method is likely vulnerable for falsely classified data with high confidence. 
This is called a \textit{label switching} problem and is an important issue to be addressed in domain adaptation. 
With label switching, even though the embedded features exhibit well-aligned clusters, the classification accuracy may not increase. 
We speculate that preventing the significant movement of the features along the data manifold can mitigate the number of misclassified data. 
It can be implemented with the manifold attainable using GANs and by penalizing the change in features when the data moves along the manifold slightly, as suggested in \cite{lgan,manifold_semisup}. 
In our opinion, this constraint would complement our proposed method and this will be investigated in our future work.




\bibliography{main}
\bibliographystyle{icml2019}

\end{document}